\documentclass{article}

\PassOptionsToPackage{numbers, compress, sort}{natbib}

\usepackage[preprint]{neurips_2025}

\usepackage[utf8]{inputenc} 
\usepackage[T1]{fontenc}    
\usepackage{hyperref}       
\usepackage{url}            
\usepackage{booktabs}       
\usepackage{amsfonts}       
\usepackage{nicefrac}       
\usepackage{microtype}      
\usepackage{xcolor}         
\usepackage{amsmath,mathtools}
\usepackage{subcaption}
\usepackage{todonotes}
\usepackage{natbib}
\usepackage{comment}
\usepackage{floatrow}
\newfloatcommand{capbtabbox}{table}[][\FBwidth]

\usepackage{nicematrix}
\usepackage[export]{adjustbox}
\usepackage{cleveref}
\usepackage[ruled,vlined]{algorithm2e}

\crefname{algocf}{alg.}{algs.}
\Crefname{algocf}{Algorithm}{Algorithms}

\crefformat{equation}{(#2#1#3)}
\crefrangeformat{equation}{(#3#1#4) to~(#5#2#6)}
\crefmultiformat{equation}{(#2#1#3)}{ and~(#2#1#3)}{, (#2#1#3)}{ and~(#2#1#3)}

\let\originalleft\left
\let\originalright\right
\renewcommand{\left}{\mathopen{}\mathclose\bgroup\originalleft}
\renewcommand{\right}{\aftergroup\egroup\originalright}

\SetAlgoInsideSkip{smallskip}

\newcommand{\bx}{\mathbf{x}}
\newcommand{\by}{\mathbf{y}}

\newcommand{\bbR}{\mathbb{R}}

\usetikzlibrary{positioning,calc} 
\tikzset{
    shifted by/.style={to path={($(\tikztostart)!#1!90:(\tikztotarget)$) -- ($(\tikztotarget)!#1!-90:(\tikztostart)$)}}, 
    shifted by/.default=2pt,standard edge/.style={very thick,-latex}, 
    back and forth between/.style args={#1 and #2}{insert path={#1 edge[standard edge,-latex,shifted by] #2 #2 edge[standard edge,shifted by] #1}}
}

\title{Sparse Techniques for Regression in \\ Deep Gaussian Processes}

\author{%
Jonas~Latz$^{1,}$\thanks{Corresponding author: \texttt{jonas.latz@manchester.ac.uk}.} \qquad Aretha~L.~Teckentrup$^{2,3}$ \qquad Simon~Urbainczyk$^{3,4}$\\
\\$^1$University of Manchester \qquad $^2$University of Edinburgh \\ $^3$Maxwell Institute for Mathematical Sciences \qquad $^4$Heriot-Watt University
}

\begin{document}

\maketitle

\begin{abstract}
    Gaussian processes (GPs) have gained popularity as flexible machine learning models for regression and function approximation with an in-built method for uncertainty quantification. However, GPs suffer when the amount of training data is large or when the underlying function contains multi-scale features that are difficult to represent by a stationary kernel. To address the former, training of GPs with large-scale data is often performed through inducing point approximations, also known as sparse GP regression (GPR), where the size of the covariance matrices in GPR is reduced considerably through a greedy search on the data set. To aid the latter, deep GPs have gained traction
    as hierarchical models that resolve multi-scale features by combining multiple GPs. 
    Posterior inference in deep GPs requires a sampling or, more usual, a variational approximation. Variational approximations lead to large-scale stochastic, non-convex optimisation problems and the resulting approximation tends to represent uncertainty incorrectly.
    In this work, we combine variational learning with MCMC to develop a particle-based expectation-maximisation method to simultaneously find inducing points within the large-scale data (variationally) and accurately train the deep GPs (sampling-based). The result is a highly efficient and accurate methodology for deep GP training on large-scale data. We test our method on standard benchmark problems.
\end{abstract}

\section{Introduction}
\label{sec:stride_introduction}
Gaussian process regression (GPR) is a popular Bayesian methodology in supervised learning, non-parametric statistics, and function approximation, with a variety of favourable features such as tractable inference and uncertainty quantification.
The GP priors used in GPR are often parametrised by a stationary covariance kernel, which may fail to accurately represent non-stationary features of the quantity to be approximated. A possible remedy is to employ a non-stationary deep GP prior instead. Such deep GPs can be constructed by, e.g., sequentially composing a number of otherwise independent GPs.
Deep GPR retains many features of GPR, but importantly makes posterior inference much more computationally challenging. Sampling-based techniques for deep GPR require repeated GPR steps, which is infeasible if the number of observed data points is large.
As a result, an efficient strategy for performing these GPR steps is needed to obtain a viable deep GPR method. In stationary GPR, sparse GPR methods have seen great success in reducing the computational complexity, particularly for large datasets. Sparse GPR relies on a low-rank approximation of the exact GP covariance. Typically, this low-rank approximation depends on a set of inducing points, that is, a set of points that is much smaller than the original dataset, but can be regarded as representative of the entire dataset.

\begin{figure*}
    \centering
    \begin{tabular}{rl}
        \includegraphics[scale=0.82]{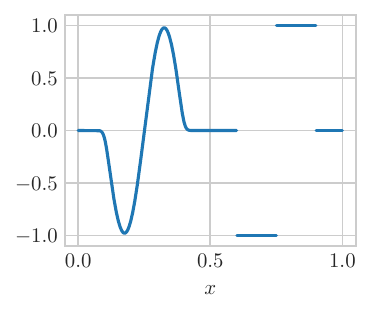} & \hspace{0.3em} \includegraphics[scale=0.82]{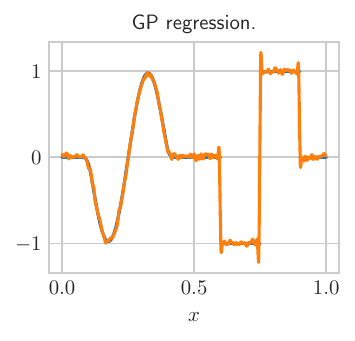} \\
        \includegraphics[scale=0.82]{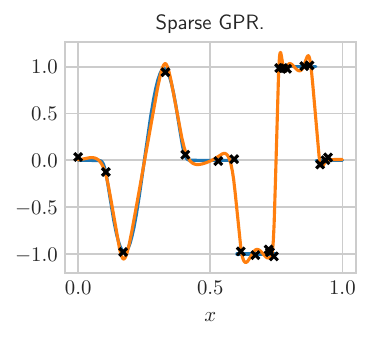} & \includegraphics[scale=0.82]{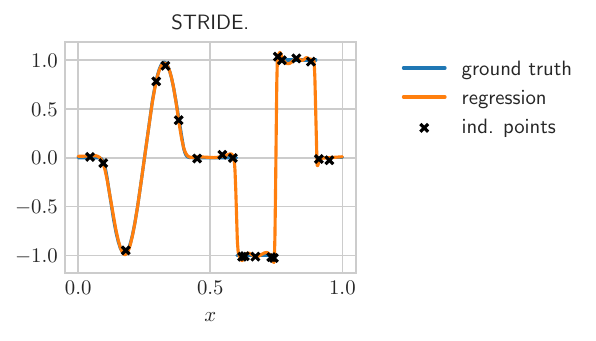}
    \end{tabular}
    \caption{Example of reconstructing a one-dimensional function using GP regression, sparse GP regression, and STRIDE, respectively.}
    \label{fig:stride_1d_example}
\end{figure*}

We illustrate the need for a methodology that combines deep and sparse GP regression techniques with a small example.
We aim to reconstruct a one-dimensional ground truth function, shown in the upper left plot in \Cref{fig:stride_1d_example}, observed at 200 locations distributed in a uniform grid on $[0, 1]$. The observations are furthermore polluted with Gaussian noise with a small variance of $0.02^2$. This ground truth is challenging in the context of GPR for the fact that it is constructed from a very smooth part on the left, and a part that contains a series of discontinuous jumps to the right. That is, we expect a stationary covariance kernel to struggle to represent this ground truth function. We show the posterior mean obtained from standard GP regression in the upper right plot in \Cref{fig:stride_1d_example}, using optimised hyper-parameters and a Gaussian covariance kernel. It is clear that we recover the left, smooth side of the ground truth very accurately, while the reconstructed function exhibits over-shooting at the jumps on the right. However, in applications involving large-scale datasets, using standard GPR is typically infeasible. Instead, a popular choice is the use of sparse GPR methods, such as described in \cite{titsias2009variational,titsias2009report}.
We employ a greedy algorithm to choose the inducing points, and optimise the hyper-parameters each time a new point is selected.
We present the result of this approach for a number of 20 inducing points in the lower left plot, with the inducing point locations marked as black crosses.
In this example, it becomes clear that the computational speed-up we gain from using sparse GPR comes at the cost of reducing the accuracy of the reconstruction. Namely, the jumps on the right-hand side of the ground truth function are recovered rather poorly by the sparse GPR mean. The goal of the method introduced in this paper is to improve the accuracy of sparse GPR, while maintaining its scalability properties. In our small example, we show the reconstruction obtained with STRIDE using a 3-layer deep GP architecture in the lower right plot in \Cref{fig:stride_1d_example}. Here, we can observe that, with the same number of inducing points, we are able to represent the ground truth more accurately. Namely, the slopes of the reconstruction around the jump locations are larger and thus more consistent with the ground truth, while slightly reducing the overshoot occurring at the same locations compared to standard and sparse GPR. We provide more details on this one-dimensional example in \Cref{sec:1d_example}.

Sparse GPR is a popular tool for solving regression problems involving large numbers of data points, see e.g.\ the works
\cite{williams2000using,smola2000sparse,csato2002sparse,titsias2009variational} as well as the unifying perspectives \cite{bui2017unifying,quinonero2005unifying}.
In STRIDE, we focus on the objective function described in \cite{titsias2009variational} to select optimal inducing points.
This is only one of many possible approaches to choosing inducing points. In \cite{bui2017unifying}, a new framework is introduced that generalises the free variational energy used in \cite{titsias2009variational} as well as the deterministic, fully independent, and partially independent training conditionals summarised in \citep{quinonero2005unifying}.
A good overview and introduction to this and other popular inducing point selection strategies in sparse GPR is given in \cite{bui2017unifying,quinonero2005unifying}.
As sparse GPR methods typically scale as $\mathcal{O}(nm^3)$ in computational complexity (and as $\mathcal{O}(nm)$ in memory), where $n$ is the number of data points and $m$ the number of inducing points, \citep{wilson2015kernel} introduce a method relying on Kronecker and Toeplitz algebra to improve the scaling w.r.t.\ $m$ in order to obtain an efficient method when large values of $m$ are needed to retain accuracy.

In the context of deep GPR, the posterior needs to be approximated with samples, e.g.\ through MCMC \citep{dunlop18,monterrubio2020posterior,sauer2023active,latz2025deep}, or a variational approximation \citep{damianou2013deep,hensman2014nested,dai2015variational,cutajar2017random}. Variational approximations are popular due to their scalability,  but challenges remain as they require non-convex large-scale optimisation, tend to misrepresent uncertainty \citep{jankowiak2020parametric}, and are not yet fully understood theoretically \citep{Blei03042017}. In this work, we therefore focus on provably convergent MCMC methods for reliable uncertainty quantification. Sparse GPR techniques based on \citep{titsias2009variational} have successfully been employed in variational approximations of deep GPs \citep{damianou2013deep,damianou2011variational}. On the other hand, \cite{rossi2021sparse} introduced a framework for inferring inducing points together with hyper-parameters in a Bayesian fashion, including an extension to deep GPs.
The method we present in this work can be seen as taking a middle route: while we employ an MCMC algorithm to obtain samples of the hidden layers, we choose inducing points through an optimisation approach, resulting in an expectation-maximisation (EM) algorithm \citep{dempster1977maximum,sundberg1974maximum,Ng2012,wei1990monte}.
This methodology can also be linked to the work of \cite{johnston2025interacting,kuntz23a} who combined Langevin sampling and gradient flows in a similar context.
Other works go beyond more classical GP tools by using a stochastic differential equation to sample inducing points \citep{xu2024sparse}, or a deep neural network to model non-stationarity in a GP \citep{wilson2016deep}.

\begin{figure*}[htb]
    \centering%
    \begin{tikzpicture}[boot/.style={thick,align=center,label={[label distance=-0.25cm]below:#1}}]
         \node[boot=$f_0$] (U0) {\includegraphics[width=2.2cm]{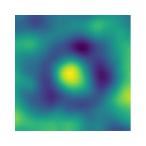}};
         \node[right=2cm of U0, boot=$f_1$] (U1) {\includegraphics[width=2.2cm]{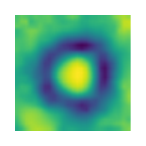}};
         \node[right=2cm of U1, boot=$f_2$] (Ul) {\includegraphics[width=2.2cm]{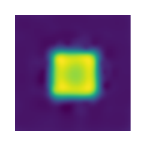}};
         \draw (U0) edge[standard edge] (U1) (U1) edge[standard edge] (Ul);
         \node[left = 0.2cm of U0.west] (conv) {\rotatebox{90}{Convolution}};
    \end{tikzpicture}%
\par\vspace{0.5cm}%
    \begin{tikzpicture}[boot/.style={thick,align=center,label={[label distance=-0.25cm]below:#1}}]
        \node[boot=$(f_0)^1$] (U00) {\includegraphics[width=2.2cm]{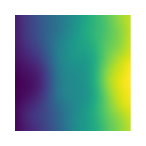}};
        \node[right=2cm of U00, boot=$(f_1)^1$] (U10) {\includegraphics[width=2.2cm]{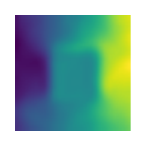}};
         
        \node[below=0.25cm of U00, boot=$(f_0)^2$] (U0d) {\includegraphics[width=2.2cm]{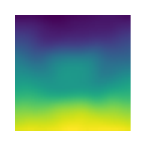}};
        \node[below=0.25cm of U10, right=2cm of U0d, boot=$(f_1)^{2}$] (U1d) {\includegraphics[width=2.2cm]{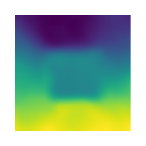}};
        
        \path (U10.east) -- (U1d.east) node[midway] (mid1) {};
        \node[right=2cm of mid1.west, align=center, boot=$f_2$] (Ul) {\includegraphics[width=2.2cm]{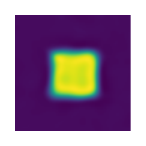}};
         
        \draw (U00) edge[standard edge] (U10) (U10.east) edge[standard edge] (Ul.west);
        \draw (U0d) edge[standard edge] (U1d) (U1d.east) edge[standard edge] (Ul.west);
        \draw (U00.east) edge[standard edge] (U1d.west);
        \draw (U0d.east) edge[standard edge] (U10.west);
        
        \node[left = 0cm of U00.west] (comp0) {\phantom{\rotatebox{90}{Composition}}};
        \node[left = 0cm of U0d.west] (compd) {\phantom{\rotatebox{90}{Composition}}};
        \path (comp0) -- (compd) node[midway] (compmid) {\rotatebox{90}{Composition}};
    \end{tikzpicture}%
    \caption{Illustration of 3-layer deep Gaussian processes with convolution (top) and composition (bottom) architectures. The input dimension is $d=2$. The deep GPs have been conditioned with respect to data from a simple piecewise-constant target.}
    \label{fig:stride_dgp_constructions}
\end{figure*}

Deep GPs can be constructed in a variety of ways, typically based on a hierarchical approach to non-stationary GP modelling \cite{dunlop18,sauer2023non}. When combining the individual GP layers in said hierarchical fashion, the most prominent example is the composition architecture as described in the seminal papers \cite{damianou2013deep,duvenaud2014avoiding}. The convolution architecture, on the other hand, uses the individual GP layers to define a non-stationary correlation length for the next layer \cite{roininen2019hyper,dunlop18}.
We illustrate both architectures in \Cref{fig:stride_dgp_constructions}. A synthesis of composition and convolution architectures was recently proposed in \cite{de2023thin}.
Another notable way of constructing deep GPs is the use of non-stationary covariance operators that are defined via the realisations of the next-lower layers \citep{latz2025deep,dunlop18}.

In the present work, we propose STRIDE (short for \emph{Sparse Techniques for Regression in Deep Gaussian Processes}), a method that combines Markov chain Monte Carlo (MCMC) with sparse GPR for an accurate and computationally efficient deep GPR. Indeed, we
\begin{itemize}
    \item[(i)] propose a sparse-Gaussian-process-marginalised MCMC sampler for deep Gaussian process regression,
    \item[(ii)] propose a heuristic greedy algorithm to find inducing points by optimising a deep variational lower bound,
    \item[(iii)] construct STRIDE by combining \textit{(i)}, \textit{(ii)} in a Monte Carlo Expectation-Maximisation methodology,
    \item[(iv)] discuss approximate sampling techniques on lower deep Gaussian process layers to reduce the overall complexity to be linear in the size of the training dataset,
    \item[(v)] verify our methodology in usual benchmark problems and in a large-scale Fashion-MNIST problem, where it beats comparable sparse Gaussian processes.
\end{itemize}

\paragraph{Structure.}
Following this introduction, we collect the technical background on conventional, deep, and sparse GPR in \Cref{Sec_GPR}. Based on this, we combine deep and sparse GP techniques to derive STRIDE in \Cref{sec:stride}. We describe the behaviour and performance of our algorithm in a range of numerical experiments, presenting their setup and results in \Cref{Sec_Exp}. Finally, we conclude in \Cref{Sec_conclusions}.

\section{Gaussian process regression and extensions} \label{Sec_GPR}
In the following, we consider a supervised learning problem: we are given data pairs $(\bx_i, y_i)_{i=1}^n \in (\mathbb{R}^d \times \mathbb{R})^n$ through $f^\dagger(\bx_i) + \varepsilon_i = y_i$, where $f^\dagger$ is an unknown function and $\varepsilon_1,...,\varepsilon_n \sim \mathrm{N}(0,\gamma^2)$ is observational noise assumed to be independent and identically distributed. We denote the full dataset divided into observation locations and values $X := (\bx_i)_{i=1}^n$ and $\by := (y_i)_{i=1}^n$. Our goal is to recover the function $f^\dagger$ from the observational data. GPR is a natural Bayesian strategy to approach problems of this form. Given a prior $\mathcal{GP}(0,k
)$, we obtain the posterior $\mathcal{GP}(\mu^\by
,k^\by
)$, with
\begin{align}
    \label{eq_GPR}
    \mu^\by
    &= k(\cdot, X)\left(k(X, X)+\gamma^2\mathrm{Id}_n\right)^{-1}\by \: , \\
    k^\by
    &= k
    - k(\cdot, X)\left(k(X, X)+\gamma^2\mathrm{Id}_n\right)^{-1}k(X,\cdot) \: . \nonumber
\end{align}
In the above, we slightly abuse notation by using $X$ as an argument of the GP covariance kernel $k$. We use $k(X, X)$ to denote the matrix resulting from evaluating $k(\bx_i,\bx_j)$ for each pair of indices $1 \leq i,j \leq n$. The expressions $k(\cdot, X)$ and $k(X, \cdot)$ are maps from $\bbR^d$ into $\bbR^n$ and contain entries $k(\cdot, \bx_i)$ for $1 \leq i \leq n$, given as a row vector and a column vector, respectively.

In the following, we discuss two extensions of GPR: deep GPR in cases where GPs are not sufficiently flexible (Subsections~\ref{sec:deep_gps}, \ref{sec:deep_gpr}) and sparse GPR as a computationally efficient approximation of GPR (Subsection~\ref{subsec_SparseGPR}).

\subsection{Deep Gaussian processes}
\label{sec:deep_gps}
A sequence of random functions $(f_\ell)_{0 \leq \ell \leq \ L}$ is called a \emph{deep Gaussian process} with $L+1$ layers if $f_0$ is a Gaussian process, and if $f_\ell$ is conditionally Gaussian for $1 \leq \ell \leq L$, i.e.
\begin{align*}
    \pi_{f_\ell| f_{\ell - 1} = u_{\ell - 1}}
    = \mathcal{GP}(0, k_\ell(\cdot, \cdot \, | \, u_{\ell - 1})) \: ,
\end{align*}
for any realisation $u_{\ell - 1}$ of $f_{\ell -1}$.

The properties of a deep GP strongly depend on the choice of the conditional covariance kernel $k_\ell$, which determines its architecture.
We consider two approaches here, namely, the convolution architecture described in \citep{paciorek03,dunlop18}, as well as the composition/warping architecture \citep{damianou2013deep,dunlop18,sauer2023active}. We illustrate both constructions in \Cref{fig:stride_dgp_constructions}, which we refer to in what follows. In this figure, we show the estimated posterior mean for each layer, conditioned on data from a piece-wise constant ground truth function.

\paragraph{Convolution.}
Covariance kernels are often constructed in the following way: we define an isotropic correlation function $\rho:\mathbb{R}\rightarrow\mathbb{R}$ and then obtain a covariance kernel through $k(\bx, \bx') = \sigma^2 \rho(\|\bx-\bx'\| \: / \: \lambda)$ for some marginal variance $\sigma^2>0$ and correlation length $\lambda > 0$. The correlation length determines the strength of long-range correlation in a GP and, thus, the oscillatory behaviour of its samples. In the convolution architecture, we now use a spatially-dependent value for $\lambda$ that is given through the realisation $u_{\ell-1}$ of the next-lower deep GP layer $f_{\ell - 1}$. A spatially variable correlation length allows us to describe multi-scale behaviour in a function, e.g., in images.
We usually set $\lambda(\bx) \coloneqq h(u_{\ell - 1}(\bx))$, where $h: \: \mathbb{R} \rightarrow \mathbb{R}_+$ is some non-negative transformation of the hidden layer values $u_{\ell - 1}$.
More precisely, in a multi-layer setting, we give the covariance function $k_\ell$ as,
\begin{align*}
    k_\ell^\mathrm{conv}&(\bx, \bx' \, | \, u_{\ell - 1})
    = \sigma^2 \, \phi(\bx, \bx') \, \rho\left(\frac{2 \, \|\bx - \bx'\|}{h(u_{\ell-1}(\bx)) + h(u_{\ell-1}(\bx'))}\right),
\end{align*}
    where
\begin{align*}
    \phi(\bx, \bx')
    = \frac{2^{d/2} \, h(u_{\ell-1}(\bx))^{d/4} \,h(u_{\ell-1}(\bx'))^{d/4}}{(h(u_{\ell-1}(\bx)) + h(u_{\ell-1}(\bx')))^{d/2}},
    \qquad
    h(z)
    = \min(\max(z, u^\mathrm{min}), u^\mathrm{max}) ^ 2\: .
\end{align*}
In the above, the pre-factor $\phi(\bx, \bx')$ ensures that the resulting covariance function remains positive semi-definite \citep{paciorek03}. The scalar transformation $h:\mathbb{R}\rightarrow\mathbb{R}^+$ given here is an example that worked well in our experiments, with $u^\mathrm{min}, u^\mathrm{max}$ as tunable parameters. Other functions that result in positive correlation length values would also be valid. Note that the correlation length can in principle be chosen to be matrix-valued, but we restrict ourselves to scalars in this work, see \cite{paciorek03,dunlop18}.
In \Cref{fig:stride_dgp_constructions}, this construction has the effect that we observe small mean values of the hidden layers close to the jump locations of the ground truth. Due to our choice of $h$, the small hidden layer values translate into a small local length scale in these areas.

\paragraph{Composition.}
The composition architecture tries to identify a transformation of the input space that best enables the top layer to model the data. That is, the layer realisations are now given by functions $u_\ell$ that map $\mathbb{R}^{d_\ell}$ into $\mathbb{R}^{d_{\ell+1}}$, where $d_0=d$ and $d_{L+1}=1$. Starting with a stationary kernel $k$, the generally non-stationary kernel of $f_\ell$ is then given by,
\begin{align*}
    k_\ell^\mathrm{comp}(\bx, \, \bx' \, | \, u_{\ell - 1})
    &= k(u_{\ell - 1}(\bx), \, u_{\ell - 1}(\bx')) \: .
\end{align*}
In this case, some of the layers may be multi-output GPs. This is also the case in our example in \Cref{fig:stride_dgp_constructions}, where we choose GPs with two-dimensional output as the hidden layers. The output values of one hidden layer serve as the spatial input of the next. The local length scale is encoded in the slope of the next-lower layer. In this example, this means that we obtain large slopes in the hidden layer mean values in areas where we require a small length scale to accurately recover the jumps of the ground truth function.

One drawback of the composition construction is that it can suffer from \emph{mode collapse}. That is, when a layer $u_{\ell-1}$ is not injective, distinct spatial locations $\mathbf{x}$, $\mathbf{x}'$ can be mapped to the same value $u_{\ell-1}(\mathbf{x}) = u_{\ell-1}(\mathbf{x}')$, meaning that these points are fully correlated in the remainder of the deep GP layers. This issue is quite prominent in one dimension, and has also been described theoretically \citep{duvenaud2014avoiding}. As a way to avoid mode collapse in this setting, we adapt the composition construction to enforce strictly monotonous layer inputs. We define this injective non-stationary kernel as,
\begin{align}
    \label{eq:inj_construction}
    k_\ell^\mathrm{inj}(x, x' \,|\, u_{\ell-1})
    &= k(g(u_{\ell-1}; x), g(u_{\ell-1}; x'))
    \intertext{for $x \in [0,1]$, where}
    g(u_{\ell-1}; x)
    &= \int_0^x \max\left(u_{\ell-1}(z), u^\mathrm{min}\right)^2 \, \mathrm{d}z \: . \nonumber
\end{align}
See also \cite{barnett2025monotonic} for a similar construction. Clearly, this setup only works in one dimension. Through the choice of $u^\mathrm{min} > 0$, we can control the maximum correlation length that can occur in the deep GP construction. Effectively, it ensures that $g(u_{\ell-1}; x) \geq (u^\mathrm{min})^2 \, x$, which puts a constraint on the minimum positive slope of the warping $g$.

\subsection{Deep Gaussian process regression}
\label{sec:deep_gpr}
The posterior in deep GPR is commonly approximated with samples, e.g.\ through MCMC \citep{dunlop18,monterrubio2020posterior,sauer2023active,latz2025deep}.
Here we especially note a helpful trick in MCMC for deep GPR, which we refer to as marginalising the outer layer. Bayes' theorem implies that
\begin{align*}
    \pi_{f_L,\ldots,f_0|\by}
    = \frac{\pi_{\by|f_L} \pi_{f_L,\ldots,f_0}}{\pi_{\by}}
    = \underbrace{\frac{\pi_{\by|f_L} \pi_{f_L|f_{L-1:0}}}{\pi_{\by|f_{L-1:0}}}}_{(*)} \underbrace{\frac{\pi_{\by|f_{L-1:0}}\pi_{f_{L-1:0}}}{\pi_{\by}}}_{(**)} \: ,
\end{align*}
where $u_{L-1:0} := u_{L-1},\ldots,u_0$ and
\begin{itemize}
    \item[$(*)$] is the conditional outer layer posterior $\pi_{f_L|\by,f_{L-1:0}}$ and given through \eqref{eq_GPR},
    \item[$(**)$] is the inner layer's posterior $\pi_{f_{L-1:0}|\by}$ that can be sampled with MCMC.
\end{itemize}
Hence, the MCMC method needs to approximate the posterior distributions on the lower layers $f_{L-1:0}$ only, but not the outer layer $f_L$.

We gain two advantages from marginalising the top layer.
Firstly, one typically wants to avoid sampling where possible, as sampling tends to be computationally expensive. Crucially, the computational cost of most sampling methods increases with the dimension of the sampled quantity.
However, there is another, more subtle benefit to treat the top layer separately. Typically, this top layer's likelihood is particularly singular (concentrated) compared to the hidden layers. This means that sampling the top layer is more costly than sampling a hidden layer, as the associated Markov chain takes longer to reach stationarity. This is also what we observe in practice. If we sample the top layer, we need a prohibitively large number of samples to obtain a reasonable approximation to the top layer mean.

\subsection{Sparse Gaussian process regression}
\label{subsec_SparseGPR}
In GPR, the inversion of the kernel matrix becomes a computational challenge for large numbers of observations $n$. Sparse GPR reduces the cost of inverting this matrix by representing the observed data through a much smaller number of \emph{virtual} observations, or inducing variables.
Here, the term virtual relates to the fact that generally, the inducing points do not have to be real observations, but can just be chosen to represent the observations well. However, in what follows, we only consider the case where the inducing points are selected from the observations.
We denote the number of inducing variables by $m$, and generally choose $m \ll n$. We then construct an approximate posterior $\pi^{\mathcal{M}}_{f|\by}$ that only requires us to work with $n\times m$ and smaller matrices. 

Suppose we have chosen a subset of the indices of the observational data points, $\mathcal{M} \subset [n]$ with $|\mathcal{M}| = m$, to represent our inducing points. Here, $[n]$ denotes the set $[n]=\{i \in \mathbb{N} \: | \: i \leq n\}$.
Let $\psi:[m] \rightarrow [n]$ denote a mapping that assigns each observation selected as an inducing point its corresponding index in the data $(\bx_i,y_i)_{i=1}^n$. Using this mapping, define the matrices $K_{mm} \in \mathbb{R}^{m \times m}$ and $K_{mn} \in \mathbb{R}^{m \times n}$, as well as the vector of covariances $K_m(\bx) \in \mathbb{R}^m$, element-wise as

\begin{align*}
    (K_{mm})_{j,j'} = k\left(\bx_{\psi(j)}, \bx_{\psi(j')}\right), \qquad
    (K_{mn})_{j,i} = k\left(\bx_{\psi(j)}, \bx_i\right), \qquad
    (K_m(\bx))_j = k\left(\bx_{\psi(j)}, \bx\right),
\end{align*}
where $i \in [n], j,j' \in [m], \bx \in \mathbb{R}^d.$

In selecting the inducing points from the observations, we follow \cite{titsias2009variational}.
We choose inducing points represented by indices $\mathcal{M}$ that minimise the KL divergence between an approximate posterior $\pi^{\mathcal{M}}_{f|\by}$ and the posterior $\mathcal{GP}(\mu^\by, k^\by)$.
As is shown in \cite{titsias2009variational,titsias2009report}, this KL divergence can be minimised by maximising the variational lower bound,
\begin{align}
    \begin{split}
        F_V(\mathcal{M})
        &\propto - \frac{1}{2} \by^T (K_{mn}^T K_{mm}^{-1} K_{mn} + \gamma^2 \mathrm{Id}_n)^{-1} \by
        - \frac{1}{2} \log\det(K_{mn}^T K_{mm}^{-1} K_{mn} + \gamma^2 \mathrm{Id}_n) \\
        &\mathrel{\phantom{\propto}}- \frac{1}{2\gamma^2}\mathrm{trace}(K_{rr} - K_{mr}^T K_{mm}^{-1} K_{mr}) \: .
    \end{split}
    \label{eq:criterion}
\end{align}
In the above, the subscript $r$ used for matrices $K_{rr}$ and $K_{mr}$ denotes the observational data indices that are not included as inducing points. That is,
\begin{align*}
    K_{rr} = \left(k\left(\bx_i, \bx_j\right)\right)_{i \in [n] \backslash \mathcal{M}, \, j \in [n] \backslash \mathcal{M}}, \qquad
    K_{mr} = \left(k\left(\bx_i, \bx_j\right)\right)_{i \in \mathcal{M}, j \in [n] \backslash \mathcal{M}} \: .
\end{align*}
The approximate posterior $\pi^{\mathcal{M}}_{f|\by} = \mathcal{GP}(\mu_\mathcal{M},k_\mathcal{M})$ obtained with sparse GPR is then given by
\begin{align*}
    \mu_\mathcal{M}(\bx)
    &= \gamma ^ {-2} K_m(\bx)^T \left(K_{mm} + \gamma ^ {-2} K_{mn} K_{mn}^T\right)^{-1} K_{mn} \by \: , \\
    k_\mathcal{M}(\bx, \bx')
    &= k(\bx, \bx') - K_m(\bx)^T K_{mm}^{-1} K_m(\bx')
    + K_m(\bx)^T \left(K_{mm} + \gamma ^ {-2} K_{mn} K_{mn}^T\right)^{-1} K_m(\bx') \: .
\end{align*}
Note that these formulas only use the inverse of $m \times m$ matrices, as opposed to $n \times n$ for standard GPR. Likewise, the memory needed to store the covariance matrices is reduced to $\mathcal{O}(mn)$.

\section{Inducing points for deep Gaussian process regression}
\label{sec:stride}
When using the marginalisation strategy in Subsection~\ref{sec:deep_gpr}, we need to repeatedly evaluate $\pi_{\by|f_{L-1},\ldots,f_0}$. Each of the latter evaluations requires a GPR step, which is prohibitively expensive if $n$ is large. In the following, we introduce an iterative procedure that allows us find a set of inducing points $\mathcal{M}$ that help us to construct approximate distributions $$\pi_{f_L|\by, f_{L-1:0}}^{\mathcal{M}}, \quad \pi_{f_{L-1:0}|\by}^{\mathcal{M}}, \quad \pi_{\by|f_{L-1:0}}^{\mathcal{M}}$$ through sparse GPR and, thus, enable deep GPR under large-scale data.

In particular, our goal is in the following to find a set of inducing points $\mathcal{M}$ that maximises
\begin{align}
    \label{eq:em_objective}
        F(\mathcal{M})
        = \int F_V(\mathcal{M}; u_{L-1:0}) \: \mathrm{d}\pi_{f_{L-1:0}|\by}^{\mathcal{M}}(u_{L-1:0}) \: ,
\end{align}
where $F_V(\mathcal{M}; u_{L-1:0})$ is given as in \eqref{eq:criterion}, but with the covariance kernel $k := k_L$ depending on lower layers $u_{L-1:0}$. To maximise $F$, we propose a Monte Carlo Expectation-Maximisation (MCEM) method, see \citep{wei1990monte} for details.

Expectation-Maximisation (EM) methods are used to maximise an objective that is typically parametrised by missing data. In an expectation step, we infer the missing data given an estimate of the optimum. In a maximisation step, this inferred missing data is then used to approximate the objective and find the next estimate of the optimum through optimisation. Both steps are re-iterated alternately until a convergence criterion is satisfied.
In our case, the objective $F$ depends on the distribution represented via the density $\pi_{f_{L-1:0}|\by}^{\mathcal{M}}$. Thus, in the expectation step, we aim at estimating the expected value as given in \cref{eq:em_objective} for a given set of inducing points $\mathcal{M}$. In the maximisation step, we optimise the approximate objective obtained in the expectation step.
However, due to the nature of our problem, we are not able to evaluate $F(\mathcal{M})$ exactly. In particular, we rely on MCMC to sample from $\pi_{f_{L-1:0}|\by}^{\mathcal{M}}$ and obtain an empirical approximation of $F(\mathcal{M})$ in the expectation step, making our method an MCEM algorithm.

Let $P_{\mathcal{M}}$ be an MCMC kernel, i.e., a Markov kernel that is stationary with respect to $\pi_{f_{L-1:0}|\by}^{\mathcal{M}}$ and ergodic. We then construct a sequence of particles and inducing point sets $\left(f_{L-1:0}^{(1,t)},\ldots,f_{L-1:0}^{(S,t)},\mathcal{M}_t\right)_{t \geq 0}$, where $S$ denotes the number of particles and $t$ denotes time. Particles and sets are given through the following two-step iteration.
\begin{itemize}
    \item[\bf (E)] Given inducing points $\mathcal{M}_t$, sample $$f_{L-1:0}^{(s,t+1)} \sim P_{\mathcal{M}_t}\left(\cdot \,\Big|\, f_{L-1:0}^{(s,t)}\right) \qquad  (s \in [S]).$$
    \item[\bf (M)] Given particles $\left(f_{L-1:0}^{(s,t+1)}\right)_{s=1}^S$, update the inducing points through
    \begin{align*}
        \mathcal{M}_{t+1} \in \mathrm{argmax}_{\mathcal{M}} \frac{1}{S} \sum_{s=1}^S F_V\left(\mathcal{M}; f_{L-1:0}^{(s,t+1)}\right) \: .
    \end{align*}
\end{itemize}
We refer to the above method as \emph{STRIDE}.
Here, \textbf{(E)} corresponds to the expectation step, and the \textbf{(M)} to the maximisation step in this MCEM algorithm, respectively.
We terminate the method at a large $T \in \mathbb{N}$. The final model is then given by
\begin{align}
    \label{eq:final_model}
    \frac{1}{S} \sum_{s=1}^S \mathcal{GP}\left(\mu_{\mathcal{M}_T}\left(\cdot \,\Big|\, f_{L-1:0}^{(s,T)}\right),k_{\mathcal{M}_T}\left(\cdot,\cdot \,\Big|\, f_{L-1:0}^{(s,T)}\right)\right) \: .
\end{align}
We summarise our algorithm in \Cref{alg:em}.
Both the expectation and the maximisation step will be described in more detail in the following two subsections. Following this, we discuss computationally efficient strategies on the lower layers.

\begin{algorithm}[ht]
    \DontPrintSemicolon
    \SetKwInput{Input}{Input}
    \SetKwInput{Output}{Output}
    \Input{Number of particles $S$, maximum time value $T$, initial hidden parameters $f_{L-1:0}^{(s,0)}$, observations $(X, \by)$, noise level $\gamma^2$.}
    Select initial inducing points $\mathcal{M}_0 \in \mathrm{argmax}_{\mathcal{M}} \frac{1}{S} \sum_{s=1}^S F_V\left(\mathcal{M}; f_{L-1:0}^{(s,0)}\right)$. \;
   
    \For{$t=0,~1,~\ldots,~T-1$}{
        \textbf{(E)}\phantom{\textbf{(M)}}Sample $f_{L-1:0}^{(s,t+1)} \sim P_{\mathcal{M}_t}\left(\cdot \,\Big|\, f_{L-1:0}^{(s,t)}\right) \quad (s \in [S])$. \;
        \textbf{(M)}\phantom{\textbf{(E)}}Update inducing points to obtain $\mathcal{M}_{t+1}$. \;
    }
    \Output{Samples of hidden parameters $f_{L-1:0}^{(s,T)}$, inducing points $ \mathcal{M}_T$.}
    \caption{STRIDE.}
    \label{alg:em}
\end{algorithm}

\subsection{Monte Carlo expectation step}
We first describe the procedure employed to obtain samples $\left(f_{L-1:0}^{(s,t+1)}\right)_{s=1}^S$, as required in the expectation step \textbf{(E)} of STRIDE.
Given some inducing points $\mathcal{M}_t$, we want to sample the hidden parameter from the density $\pi_{f_{L-1:0}|\by}^{\mathcal{M}_t}$. This density is given through Bayes' Theorem
\begin{align*}
    \pi_{f_{L-1:0}|\by}^{\mathcal{M}_t}
    \propto \pi_{\by|f_{L-1:0}}^{\mathcal{M}_t} \, \pi_{f_{L-1:0}} \: ,
\end{align*}
see also $(**)$. Combining the covariance structure defined by $f_{L-1:0}$ and $\mathcal{M}_t$ with the observational noise covariance, we can express the likelihood term as,
\begin{align*}
    \pi_{\by|f_{L-1:0}}^{\mathcal{M}_t}
    &\propto \frac{1}{\sqrt{\det Q_{nn}^{L-1:0}}} \: \exp\left(-\frac{1}{2} \|\by\|^2_{Q_{nn}^{L-1:0}}\right) \: ,
\intertext{where}
    \mathrel{\phantom{=}} Q_{nn}^{L-1:0}
    &= K_{mn}(f_{L-1:0})^T K_{mm}(f_{L-1:0})^{-1} K_{mn}(f_{L-1:0})
    + \gamma^2 \mathrm{Id}_n \: .
\end{align*}
The logarithmic likelihood (omitting constants) is identical to the first two terms in our definition of $F_V$ in \cref{eq:criterion}.
In order to compute samples of this density, we employ an MCMC algorithm as described in, e.g., \cite{cotter2013mcmc,dunlop18,latz2025deep,yu2011center}. In particular, we whiten the prior given by the density $\pi_{f_{L-1:0}}$ \citep{papaspiliopoulos2007general,yu2011center} and then apply the preconditioned Crank--Nicolson (pCN) proposal \citep{cotter2013mcmc}.

We note that the determinant of $Q_{nn}^{L-1:0}$ can be computed in $\mathcal{O}(m^3)$ using the Weinstein--Aronszajn identity. We usually employ multiple subsequent MCMC steps in a single expectation step to ensure that the particles approximate $\pi_{f_{L-1:0}|\by}^{\mathcal{M}_t}$ well.

\subsection{Maximisation step}
During the maximisation step \textbf{(M)} of STRIDE, we aim at finding a set of inducing points $\mathcal{M}_{t+1}$ that maximises the expected value of $F_V(\mathcal{M}; f_{L-1:0})$, which we approximate by
\begin{align*}
    \frac{1}{S} \sum_{s=1}^S F_V\left(\mathcal{M}; f_{L-1:0}^{(s,t+1)}\right) \eqqcolon F_t\left(\mathcal{M}\right) \: .
\end{align*}
Finding the global maximiser of $F_t$ likely requires an enumeration of $\{\mathcal{M} \subseteq [n]: |\mathcal{M}| = m\}$ which is infeasible. Instead, we follow a heuristic greedy strategy.
We select the initial set of inducing points $\mathcal{M}_0$ in the algorithm using a strategy motivated by \cite{titsias2009variational}: starting from an empty set, we add the observation that leads to the biggest increase of the objective. In practice, one chooses these observations from sets of candidate points $\mathcal{J}$ with cardinality $|\mathcal{J}| \ll n$ that are sampled randomly at each step to preserve computational efficiency.
Elements are added in this greedy fashion until the maximum cardinality $m$ is reached.

When updating an existing set $\mathcal{M}_t$ to obtain $\mathcal{M}_{t+1}$ in STRIDE, we proceed as follows. We loop over every element in $\mathcal{M}_t$ and measure its contribution to the overall objective value by excluding it from $\mathcal{M}_t$ and evaluating $F_t$. In this way, we determine the element that, through exclusion, \emph{decreases} $F_t$ \emph{the least}, and exclude it from $\mathcal{M}_t$. We then loop over all observations in a set of candidates $\mathcal{J} \subset [n] \backslash \mathcal{M}_t$ to determine the element that, through adding, \emph{increases} $F_t$ \emph{the most}. We then include this element in our set of inducing points. This birth/death-type update can be repeated for a number of iterations. Our approach is closely related to the idea of exchange algorithms used in optimal experimental design, such as described in \cite{nguyen1992review,mitchell2000algorithm}. We choose the resulting set of inducing points as our updated set $\mathcal{M}_{t+1}$. 
As mentioned, the set $\mathcal{J}$ is generally chosen to be of fixed cardinality $|\mathcal{J}| \ll n$ to control the cost of the update. As a result, the number of evaluations of $F_t$ needed for the update step is $R \, (m+|\mathcal{J}|)$, where $R$ is the number of update step repetitions.
We provide an overview of this maximisation step in \Cref{alg:update}.

\begin{algorithm}[ht]
    \DontPrintSemicolon
    \SetKwInput{Input}{Input}
    \SetKwInput{Output}{Output}
    \Input{Inducing points $\mathcal{M}_t$, observations $\mathcal{M}^\ast$, objective $F_t$, cardinality $J$, number of steps $R$.}
    Set $\overline{\mathcal{M}} \leftarrow \mathcal{M}_t$. \;
    
    \For{$r=0,~1,~\ldots,~R-1$}{
        \tcp{Find point with least contribution.}
        Set $\overline{F} \leftarrow - \infty$. \;
        \For{$z \in \overline{\mathcal{M}}$}{
            \If{$F_t(\overline{\mathcal{M}} \backslash \{z\}) > \overline{F}$}{
                Set $\overline{z} \leftarrow z$. \;
                Set $\overline{F} \leftarrow F_t(\overline{\mathcal{M}} \backslash \{z\})$. \;
            }
        }
        Set $\overline{\mathcal{M}} \leftarrow \overline{\mathcal{M}} \backslash \{\overline{z}\}$. \;
        \;
        \tcp{Replace with point with largest contribution.}
        Set $F^\ast = - \infty$. \;
        Randomly draw candidate points $\mathcal{J} \subset \mathcal{M}^\ast \backslash \overline{\mathcal{M}}$ with $|\mathcal{J}| = J$. \;
        \For{$z \in \mathcal{J}$}{
            \If{$F_t(\overline{\mathcal{M}} \cup \{z\}) > F^\ast$}{
                Set $z^\ast \leftarrow z$. \;
                Set $F^\ast \leftarrow F_t(\overline{\mathcal{M}} \cup \{z\})$. \;
            }
        }
        Set $\overline{\mathcal{M}} \leftarrow \overline{\mathcal{M}} \cup \{z^\ast\}$. \;
    }
    Set $\mathcal{M}_{t+1} \leftarrow \overline{\mathcal{M}}$. \;
    
    \Output{Updated inducing points $\mathcal{M}_{t+1}$.}
    \caption{Birth/death algorithm.}
    \label{alg:update}
\end{algorithm}

\subsection{Linear complexity on lower layers}
The main advantage of using sparse GPR methods such as \cite{titsias2009variational} is that they can be computed with an $\mathcal{O}(n)$ cost, as opposed to $\mathcal{O}(n^3)$ for standard GPR.
More precisely, we consider the computational cost during training (offline phase), the computational cost of inference (online phase), as well as the memory consumption here. Asymptotically, sparse GPR as described in \cite{titsias2009variational} has an online computational cost of $\mathcal{O}(nm^2)$, an offline computational cost of $\mathcal{O}(nm^3|\mathcal{J}|)$, as well as a memory cost of $\mathcal{O}(nm)$.

We aim at achieving comparable properties in STRIDE.
However, in order to be able to choose $\mathcal{J}$ as a subset of the observations, we need to evaluate the hidden layers of the deep GP at every observed location. This leads to having to construct covariance matrices of size $n \times n$, and sampling the hidden layers to obtain the covariance of the top layer then results in a computational cost of $\mathcal{O}(n^3)$, which is the complexity of the required Cholesky decomposition. We now propose two approaches for efficient computation on the lower layers.

\paragraph{Adaptive cross-approximation.} 
We can sample the hidden layers approximately using an approach that is cheaper than $\mathcal{O}(n^3)$. In our experiments, we use an adaptive cross-approximation (ACA) algorithm \citep{harbrecht2012lowrank,harbrecht2015efficient,kressner20}, which approximates the Cholesky factor of the covariance matrix.
More specifically, ACA determines a set of indices $I$ that identify the columns of the covariance matrix that can be used for an accurate low-rank approximation of the same covariance matrix.
More precisely, let $K\in\mathbb{R}^{n\times n}$ be a covariance matrix. Then, we denote by $K_{n,I}$ the matrix formed of the columns of $K$ at indices given by $I$. We furthermore denote by $K_{I,I}$ the matrix formed of the rows of $K_{n,I}$ at indices $I$. The low-rank approximation to $K$ induced by the ACA indices $I$ is then given by $K\approx K_{n,I} K_{I,I}^{-1} K_{n,I}^T$.
We employ the ACA algorithm as presented in \citep{kressner20} to compute a set of observation points, represented by indices $I$, that result in a low-rank approximation of the covariance matrix at the top layer. We use the same set of indices to approximate the covariance matrices at the hidden layers. Now, in order to obtain an approximate sample of the hidden layers, recall that our MCMC routine samples centred representations of the hidden layers, which we denote by $\xi_{L-1:0}^{(s,t)}$ here. The actual size of $\xi_{L-1:0}^{(s,t)}$ will depend on the rank $|I|$ of our approximation. We present the routine used for obtaining approximate samples $\hat{f}_{L-1:0}^{(s,t)}$ from the centred variable samples $\xi_{L-1:0}^{(s,t)}$ in \Cref{alg:aca}.

One can either select this set of indices once at the start of our EM algorithm and leave it fixed throughout, or, after each maximisation step \textbf{(M)}, recompute the set of indices to reflect that the covariance matrix we aim to approximate has changed. In our experiments using ACA, we employ this latter approach. However, note that, since we are sampling whitened variables in the expectation step \textbf{(E)}, this means that we also need to update these sampled variables to ensure they represent the same hidden parameters $f_{L-1:0}$ after updating our approximation of the covariance.

\begin{algorithm}[ht]
    \DontPrintSemicolon
    \SetKwInput{Input}{Input}
    \SetKwInput{Output}{Output}
    \Input{Centred MCMC sample $\xi_{L-1:0}^{(s,t)}$, pre-computed ACA indices $I$.}
    \For{$\ell=0, ~\ldots, ~L-1$}{
        \eIf{$\ell = 0$}{
            Set covariance operator $k_\ell$ to stationary bottom layer kernel. \;
        }
        {
            Evaluate covariance operator $k_\ell\left(\cdot,\cdot \,\Big|\, \hat{f}_{\ell-1}^{(s,t)}\right)$ w.r.t.\ lower layer approximation $\hat{f}_{\ell-1}^{(s,t)}$. \;
        }
        \;
        Use $k_\ell$ to compute covariance matrices $K_{n,I}$ and $K_{I,I}$. \;
        Compute $U$ as the upper Cholesky factor of $K_{I,I}$. \;
        Evaluate approximate hidden layer sample $\hat{f}_\ell^{(s,t)} = K_{n,I} \, U^{-1} \, \xi_\ell^{(s,t)}$. \;
    }
    \Output{Approximate hidden parameters sample $\hat{f}_{L-1:0}^{(s,t)}$.}
    \caption{Sampling with adaptive cross-approximation (ACA).}
    \label{alg:aca}
\end{algorithm}

\paragraph{Fully sparse.}
We can sample the hidden layers at the current inducing points only. In this case, sampling the hidden layers becomes very cheap. The covariance at the top layer is then based on an interpolation of the next-lower layer to all observation locations. We perform this interpolation using standard GPR.
The covariance kernel used here constitutes another set of tuning parameters. In practice, we simply use a stationary GP with the same marginal variance, correlation length and observation noise variance that the deep GP itself is initialised with.
Using this \emph{fully sparse} approach of sampling only at the inducing point locations also means that we need to update our hidden parameters every time we change the inducing points $\mathcal{M}_t$. That is, once the inducing points change, we now need evaluations of the hidden layers at these new locations. Once more, we obtain these through interpolation using GPR.

We summarise the fully sparse approach to sampling the hidden layers in \Cref{alg:sparse}. Here, we denote by $K_{m,m}$ the matrix of covariances between all points in $\mathcal{M}_t$, evaluated with the covariance kernel $k_\ell$. However, in order to be able to compute values of $F_V$, or evaluate the final deep GPR model given in \Cref{eq:final_model}, we require the last sampled layer, $\hat{f}_{L-1}^{(s,t)}$ to include function values not just at the inducing points $\mathcal{M}_t$, but at all observation locations.
As a result, we include an extra interpolation step, in which we use stationary sparse GP regression to predict values of $\hat{f}_{L-1}^{(s,t)}$ at the missing locations.

\begin{algorithm}[ht]
    \DontPrintSemicolon
    \SetKwInput{Input}{Input}
    \SetKwInput{Output}{Output}
    \Input{Centred MCMC sample $\xi_{L-1:0}^{(s,t)}$, set of inducing points $\mathcal{M}_t$.}
    \For{$\ell=0, ~\ldots, ~L-1$}{
        \eIf{$\ell = 0$}{
            Set covariance operator $k_\ell$ to stationary bottom layer kernel. \;
        }
        {
            Evaluate covariance operator $k_\ell\left(\cdot,\cdot \,\Big|\, \hat{f}_{\ell-1}^{(s,t)}\right)$ w.r.t.\ lower layer approximation $\hat{f}_{\ell-1}^{(s,t)}$. \;
        }
        \;
        Use $k_\ell$ to compute covariance matrix $K_{m,m}$. \;
        Compute $C$ as the lower Cholesky factor of $K_{m,m}$. \;
        Evaluate approximate hidden layer sample $\hat{f}_\ell^{(s,t)} = C \, \xi_\ell^{(s,t)}$. \;
    }
    Interpolate $\hat{f}_{L-1}^{(s,t)}$ to all observation locations using sparse GPR. \;
    \Output{Approximate hidden parameters sample $\hat{f}_{L-1:0}^{(s,t)}$.}
    \caption{Fully sparse sampling approach.}
    \label{alg:sparse}
\end{algorithm}

\subsection{Computational cost and memory consumption.}
\label{sec:cost}
Using either of these two hidden layer approximations, we can now state the asymptotic costs associated with STRIDE. In the case of using an adaptive cross-approximation for the hidden layers, we assume that the ACA rank is chosen as a linear function of $m$. The computational cost of performing inference with the trained model is then given by $\mathcal{O}(nm^2S)$. That is, the online cost scales the same as for sparse GPR, with an additional dependence on the number of particles $S$. The computational cost of running STRIDE for training gives rise to a more complicated expression.
Separated by expectation and maximisation steps, these costs are as follows,
\begin{align*}
    &\text{combined cost of expectation steps \textbf{(E)}:}
    &\qquad \mathcal{O}(STK_\mathrm{MCMC}(nm^2 + m^3 (L-1))) \: , \\
    &\text{combined cost of maximisation steps \textbf{(M)}:}
    &\qquad \mathcal{O}(nm^2STR(m+|\mathcal{J}|)) \: .
\end{align*}
Here, we denote the number of EM iterations by $T$, the number of MCMC steps used in the expectation step \textbf{(E)} by $K_\mathrm{MCMC}$, and the number of birth/death updates in the maximisation step \textbf{(M)} by $R$.
Overall, we can observe that the computational cost remains linear in $n$.

The memory required for our approach still scales as $\mathcal{O}(nm)$ only. However, this is the minimum memory consumption and would increase for more parallel implementations. E.g., when fully parallelising the maximisation step \textbf{(M)}, one requires memory scaling as $\mathcal{O}(nm|\mathcal{J}|)$.

\section{Numerical experiments}
\label{Sec_Exp}
We now test our algorithm in a number of numerical examples. In a first step, we explore the influence of the number of inducing points, $m$, and the number of layers, $L$, on the regression error. We perform this analysis in the setting of the one-dimensional toy problem already discussed in \Cref{sec:stride_introduction}.
We then turn towards usual machine learning test problems. Namely, we consider the performance in three small-scale regression problems using datasets from the UCI Machine Learning Repository \citep{boston_housing,concrete_compressive_strength_165,energy_efficiency_242}.
These are benchmark datasets that allow us to compare our work to, e.g., \cite{titsias2009variational,de2023thin,rossi2021sparse}.
Since the deep GP in our approach can be specified in a multitude of ways, we explore a range of possible setups in these examples. Mainly, we aim to investigate the effect of the deep GP construction type, as well as number of hidden layers.
In a final experiment, we test the performance of our approach on a variation of the Fashion-MNIST dataset \citep{xiao2017fashion}. Here, we consider features extracted using an auto-encoder, as presented in \citep{calder2020poisson,fashion_mnist_vae}. This dataset is much larger (70,000 points with 30 dimensions) than the UCI datasets considered. Importantly, standard GPR becomes infeasible on such large datasets, so that we have to resort to sparse GPR methods.
All experiments were performed on an AMD EPYC 7513 32-core processor; we allocated 16 GB of RAM.

\subsection{One-dimensional toy problem}
\label{sec:1d_example}
As a first example, we consider the ground truth function already mentioned in \Cref{sec:stride_introduction} and shown in the top left plot in \Cref{fig:stride_1d_example}.
A similar function was also considered in \cite{Osborne,roininen2019hyper}. In particular, the ground truth we consider here is defined on the interval $[0, 1]$ and given by,
\begin{align*}
     f^\dagger(x) &=
    \begin{cases}
        \int_{0.1}^{0.4}\frac{1}{\sqrt{2 \pi \: 10^{-4}}} \mathrm{e}^{- \frac{(x-y)^2}{2 \cdot 10^{-4}}} \sin\left(\frac{\pi (y - 0.25)}{0.15}\right) \mathrm{d}y \:,  &\text{if } 0 \leq x \leq 0.6 \:, \\
        -1 \:, \hspace{0pt - \widthof{$-1 \:,$}} \hfill &\text{if } 0.6 < x \leq 0.75 \:, \\
        1 \:, \hspace{0pt - \widthof{$1 \:,$}} \hfill &\text{if } 0.75 < x \leq 0.9 \:, \\
        0 \:, \hspace{0pt - \widthof{$0 \:,$}} \hfill &\text{if } 0.9 < x \leq 1 \:.
    \end{cases}
\end{align*}
Our goal is to use this as a simple example to compare STRIDE with standard GPR and the sparse GPR approach in \cite{titsias2009variational}, and to investigate the influence of the number of inducing points $m$ and the number of layers $L$ on the reconstruction result.
We observe the ground truth function at $n=200$ locations chosen as a uniform grid in $[0, 1]$, and pollute the observations with i.i.d.\ Gaussian noise with variance $\gamma^2 = 0.02^2$. As the ground truth contains jumps, we choose the kernel $k$ used for GP regression, sparse GPR, and to construct our deep GP prior, as the Mat\'ern kernel with smoothness $\nu=3/2$ for its low regularity \citep{williams2006gaussian}. That is,
\begin{align*}
    k(\bx, \bx')
    &= \sigma^2 \left(1 + \frac{\sqrt{3} \|\bx - \bx'\|_2}{\lambda}\right) \exp\left(- \frac{\sqrt{3} \|\bx - \bx'\|_2}{\lambda}\right) \: .
\end{align*}
Both for GPR and sparse GPR, we choose the marginal variance $\sigma^2$ and the correlation length $\lambda$ to maximise the likelihood of the observed data.

In order to define our deep GP prior to use in STRIDE, we employ a variation of the composition construction.
Namely, we use the injective non-stationary kernel defined in \cref{eq:inj_construction}, setting $u^\mathrm{min}=0.3$.
In terms of parameters for STRIDE, we choose a total of 10 EM iterations, with each expectation step \textbf{(E)} consisting of 1000 MCMC steps, and each maximisation step \textbf{(M)} consisting of 5 inducing point update steps. The candidate set in the update steps, $\mathcal{J}$, is chosen as the entire observation data. As the number of particles $S$, we choose $S = 10$.

\begin{figure}
    \includegraphics[height=0.3\linewidth]{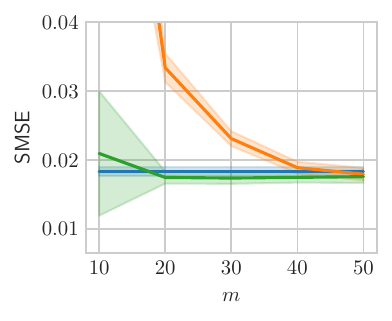}
    \quad
    \includegraphics[height=0.3\linewidth]{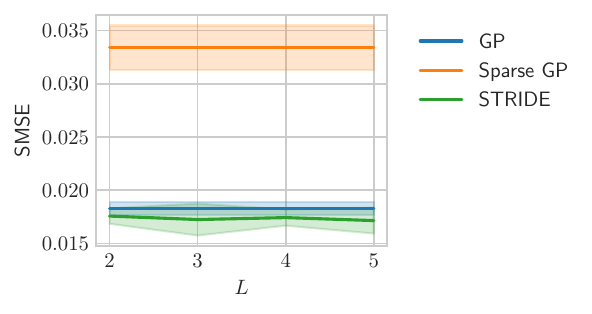}
    \caption{Comparison of errors for different values of $m$ and numbers of layers $L$ in the 1D example.}
    \label{fig:1d_comparison}
\end{figure}

First, we aim to explore the influence of the number of inducing points $m$ on the regression result. We select $m\in\{10, 20, 30, 40, 50\}$, and set the number of layers in the deep GP to $L=3$. We measure the quality of the regression results using the standardised mean square error (SMSE) as defined in \citep{williams2006gaussian}. We repeat the experiment with 5 different random seeds, that is, with 5 different realisations of the observation noise. We present the results in the plot on the left in \Cref{fig:1d_comparison}. We show lines for the mean SMSE of standard GPR (independent of $m$), sparse GPR, and STRIDE, with the shaded areas indicating a range of four ($\pm2$) standard deviations around the mean error. We can observe that for sparse GPR, we need about 40 inducing points to reach the accuracy of GPR using the full observations. STRIDE, however, has the advantage of allowing for a non-stationary prior. As a result, we only require about 20 inducing points to reach an accuracy that is slightly better than standard GPR. Adding more inducing points then does not visibly affect the outcome of STRIDE.

In order to compare different numbers of layers in our deep GP model, we now set $m=20$. Otherwise, the setup of our experiment remains identical. We furthermore choose $L\in\{2, 3, 4, 5\}$. The results of this experiment are presented in the plot on the right in \Cref{fig:1d_comparison}. Again, we show lines for the mean SMSE of standard GPR (independent of $L$), sparse GPR (independent of $L$), and STRIDE, with the shaded areas representing a range of four standard deviations around the mean. We observe that the SMSE values obtained through STRIDE remain almost constant w.r.t.\ the number of layers $L$. That is, while adding one hidden layer seems to offer a slight advantage over standard GPR due to implementing a non-stationary prior, adding more layers does not seem improve the outcome in this example. This aligns well with the findings of \cite{dunlop18} in their one-dimensional numerical experiment investigating the number of layers needed to accurately recover a non-smooth ground truth function.

\subsection{Standard UCI datasets}
\label{sec:uci}
For our first experiment, we now consider the energy efficiency, the concrete compressive strength, and the Boston housing datasets from the UCI Machine Learning Repository. These are relatively small, with numbers of data points ranging from 506 to 992. As a result, we can evaluate standard GPR on these datasets, and use the results for comparison.
As we are using their strategy for selecting inducing points, we also compare with the sparse GPR approach from \citep{titsias2009variational}.
Furthermore, the datasets are small enough to run deep GPR without making use of inducing points. We refer to this as \emph{standard} deep GPR. Both standard GPR and standard deep GPR would quickly become infeasible for larger datasets, since their computational cost grows cubically in $n$.

The goal of our experiment is to get an overview of the effect of choosing different architectures and approximation methods of the hidden layers. Our setup is as follows. The kernel $k$ is chosen as the Gaussian kernel, also known as the squared exponential or RBF kernel,
\begin{align*}
    k(\bx, \bx')
    &= \sigma ^ 2 \: \exp\left(- \frac{\|\bx - \bx'\|_2 ^ 2}{2 \lambda^2}\right) \: .
\end{align*}
The marginal variance $\sigma^2$, correlation length $\lambda$, and observation noise variance $\gamma^2$ are estimated through likelihood maximisation for the standard GP, and through maximising $F_V$ as given in \cref{eq:criterion} for the sparse GP approach.
The optimal hyper-parameters found in sparse GPR are then used to set up our deep GP model. That is, we use the same noise variance, and set the marginal variance of the top layer to the marginal variance obtained through optimisation. The marginal variances of the hidden layers are set to 1. The hidden layers $f^{(s,0)}_{L-1:0}$ are initialised so that the resulting covariance of the top layer is stationary with the correlation length found through hyper-parameter optimisation for each of the particles, $s=1, \ldots, S$.

For each model relying on inducing points, we select $m$ as 50. We run STRIDE for 10 EM iterations, each expectation step \textbf{(E)} consisting of 400 MCMC steps, and each maximisation step \textbf{(M)} consisting of 5 inducing point update steps. As a maximum size of the candidate set in the update steps, $|\mathcal{J}|$, we choose 500. We simulate $S=50$ different particles each time.

\begin{figure*}
    \centering
    \begin{subfigure}[t]{0.355\textwidth}
        \raggedleft
        \includegraphics[height=4.5cm]{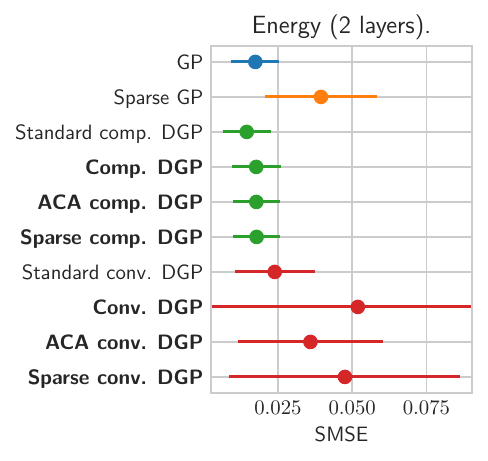}
    \end{subfigure}%
    \begin{subfigure}[t]{0.215\textwidth}
        \raggedleft
        \includegraphics[height=4.5cm]{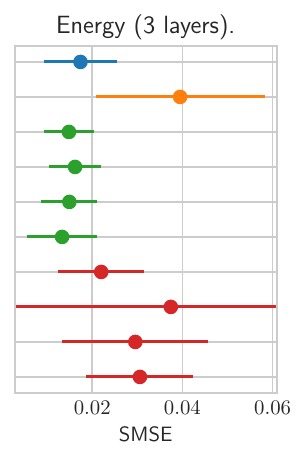}
    \end{subfigure}%
    \begin{subfigure}[t]{0.215\textwidth}
        \raggedleft
        \includegraphics[height=4.5cm]{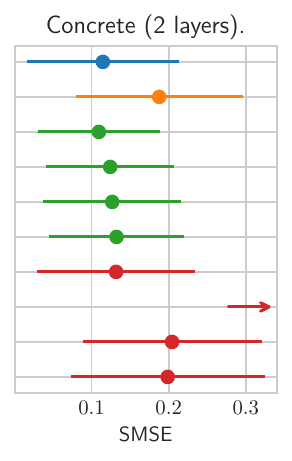}
    \end{subfigure}%
        \begin{subfigure}[t]{0.215\textwidth}
        \raggedleft
        \includegraphics[height=4.5cm]{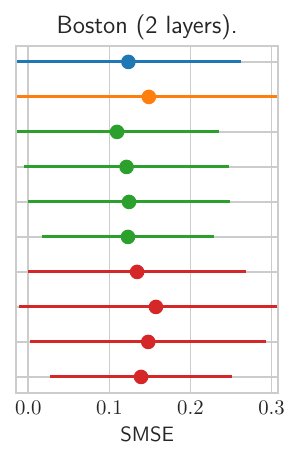}
    \end{subfigure}%
    \caption{Overview of obtained SMSE values in ten-fold cross-validation for three UCI datasets.}
    \label{fig:methods}
\end{figure*}

We use ten-fold cross-validation on each of the datasets to evaluate the SMSE values of our individual GP and deep GP setups.
The results are shown in \Cref{fig:methods}. In the left two plots, we obtain the SMSE for the energy efficiency dataset, using two and three layers in the deep GP models, respectively. In the third plot from the left, we show SMSE values for the concrete compressive strength dataset and in the plot on the right we consider the Boston housing dataset. For these last two plots, we use two layers in the deep GP models. We give the mean of the error values over the ten training/test data splits, as well as error bars of twice the standard deviation. The deep GP architectures are given in abbreviated form (\emph{comp.} and \emph{conv.}). The use of our two approaches to approximating hidden layers is marked by either \emph{ACA} or \emph{sparse}. The setups that use STRIDE to perform regression are highlighted with a bold font.

We observe that for all three datasets, using full GPR gives better results than sparse GPR, which is expected. Using a deep GP, we obtain similar, or sometimes better performance than the standard GP. Generally, the standard deep GP performs best for each architecture, giving an indication of how much using a non-stationary model could improve the error on the dataset in question. The setups using an inducing points approximation on the top layer then perform similarly.
However, we note that there is a large variance in the reconstruction errors when using the convolution architecture when sampling the full hidden layers. In particular, this setup fails to satisfyingly perform regression on the concrete dataset, leading to errors larger than 1 (as indicated by an arrow here). Approximating the hidden layers (using ACA or the fully sparse approach) appears to have a regularising effect when using the convolution architecture, as the SMSE values for these approaches are much closer to the standard deep GP values.

We use two-layer deep GPs in this experiment for the reason that the UCI datasets given here are known to be represented well by stationary Gaussian processes. As a result, we do not expect that adding more layers would improve the reconstruction results here. In fact, this is confirmed by our observation that the errors obtained for the energy efficiency dataset using two layers and three layers, respectively, appear very similar.

Generally speaking, the standard deviations shown around the mean errors are especially large in the plot concerning the Boston housing dataset, as the regression results varied significantly between different splits of the data.
As a result, while the mean of the obtained errors makes it appear that the deep GP models represent the data more accurately than the stationary GP models, we cannot come to any conclusions here due to the variance of our results.
When comparing the two different deep GP constructions, the composition architecture results in a better prediction error for each of the datasets. A reason could be that the convolution architecture makes use of GP layers with a one-dimensional output only, so that it is less flexible when storing information on the hidden layers compared to the composition architecture. We further note that using three instead of two layers improves the regression result for the energy efficiency dataset. The reduction in error can be substantial and can be observed for each of the deep GP setups.

\begin{figure*}
    \centering
    \begin{subfigure}[t]{0.355\textwidth}
        \raggedleft
        \includegraphics[height=4.5cm]{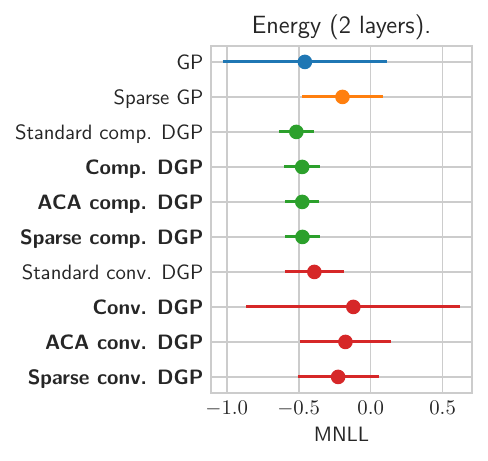}
    \end{subfigure}%
    \begin{subfigure}[t]{0.215\textwidth}
        \raggedleft
        \includegraphics[height=4.5cm]{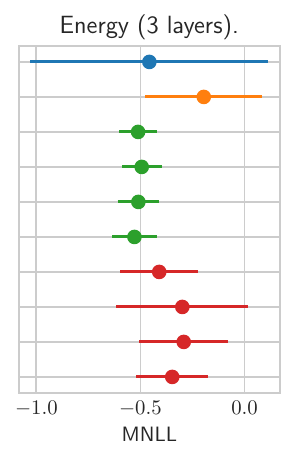}
    \end{subfigure}%
    \begin{subfigure}[t]{0.215\textwidth}
        \raggedleft
        \includegraphics[height=4.5cm]{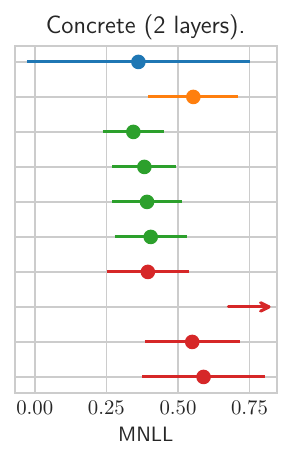}
    \end{subfigure}%
        \begin{subfigure}[t]{0.215\textwidth}
        \raggedleft
        \includegraphics[height=4.5cm]{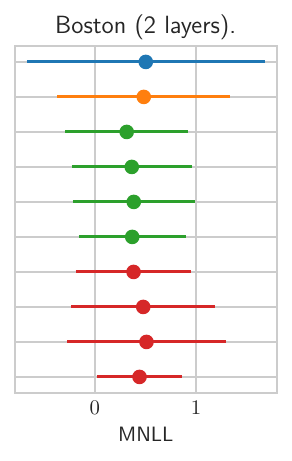}
    \end{subfigure}%
    \caption{Overview of obtained MNLL values in ten-fold cross-validation for three UCI datasets.}
    \label{fig:methods_mnll}
\end{figure*}

As an alternative to the SMSE values presented here, we also report mean negative log-likelihood (MNLL) values for these experiments.
To be precise, we compute the MNLL as,
\begin{align}
    \label{eq:mnll}
    \mathrm{MNLL}
    = \frac{1}{2} \, \log\left(2 \pi \gamma^2\right) + \frac{1}{2 p} \, \left\|\by^\mathrm{test} - \by^\mathrm{pred}\right\|^2_{\gamma^2\mathrm{Id}_p} \: .
\end{align}
Here, $\by^\mathrm{test}$ denotes the test data observations, and $\by^\mathrm{pred}$ is the prediction obtained through (potentially deep/sparse) GP regression. The integer $p$ represents the number of observations in our test data. Smaller MNLL values are generally better.
As for the SMSE, we report mean MNLL values as well as a confidence intervals of two standard deviations computed by repeating the experiment for ten different training/test data splits. The corresponding error bar plots are presented in \Cref{fig:methods_mnll}.

As is clear from the formula given in \cref{eq:mnll}, the MNLL can be seen as a rescaling of the SMSE. As a result, our interpretation of the plots in \Cref{fig:methods_mnll} is almost identical with our discussion of the SMSE results. However, the choice of $\gamma^2$ is of importance here. In order to compute the MNLL, we use the same noise variance as used in our regression model. That is, $\gamma^2$ is computed via maximum likelihood estimation for standard GPR, and through maximising $F_V$ for sparse GPR. For STRIDE, we use the same noise variance as for sparse GPR. Since the estimated value of $\gamma^2$ was generally smaller for standard GPR compared to the other approaches, we observe a larger variance in the corresponding MNLL values than in the MNLL values obtained through sparse GPR and STRIDE.

We briefly analyse the compute times for the composition architecture with two layers on the energy efficiency dataset. For a single run, using the standard deep GP took \mbox{3,116 s} of (offline) training, as well as \mbox{12.74 s} for the prediction on the test data, in what can be considered the online cost. When using STRIDE, the training time was reduced to \mbox{1,069 s}, and the prediction time went down to \mbox{0.095 s}. Using ACA and the fully sparse approach resulted in similar training times for this relatively small dataset (\mbox{1,323 s} and \mbox{1,181 s}, respectively). The prediction time for ACA was \mbox{0.093 s}, and \mbox{0.057 s} for the fully sparse approach.

\subsection{Large dataset (Fashion-MNIST)}
\label{sec:fashion}
We now turn towards performing regression on a more challenging dataset. Fashion-MNIST is a popular benchmark for machine learning methods, containing 70,000 greyscale images, each associated with a class label represented by an integer number \citep{xiao2017fashion}. As the images are of dimension 28$\times$28, the computational burden of using a deep GP with the composition architecture would make regression infeasible. However, the dimension of the data points can be reduced by techniques such as variational auto-encoders (VAEs) \citep{calder2020poisson}. In our case, we use a variant of the dataset where the data points have already been transformed by such a VAE. Using such a transformation also has the advantage that it allows us to compare the data points, which represent images in this case, in the Euclidean distance.

\begin{figure}
    \begin{floatrow}
        \ffigbox{%
            \includegraphics[height=0.525\linewidth]{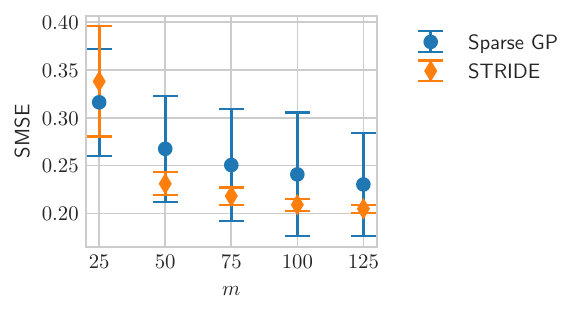}
        }{%
            \caption{Plot of SMSE values obtained for varying $m$ in the Fashion-MNIST example.}%
            \label{fig:fashion_mnist}%
        }
        \quad
        \capbtabbox{%
            \begin{tabular}{rcc}\toprule
                $m$ & Sparse GP & STRIDE \\ \midrule
                25 & 0.3162 (0.0281) & 0.3380 (0.0288) \\
                50 & 0.2676 (0.0275) & 0.2310 (0.0061) \\
                75 & 0.2506 (0.0291) & 0.2181 (0.0045) \\
                100 & 0.2408 (0.0324) & 0.2091 (0.0031) \\
                125 & 0.2303 (0.0268) & 0.2049 (0.0022) \\ \bottomrule
            \end{tabular}
            \vspace{4ex}
            }{%
                \caption{Table of SMSE values obtained for varying $m$ in the Fashion-MNIST example.}%
                \label{tab:fashion}%
        }
    \end{floatrow}
\end{figure}

We want to explore the effect of the number of inducing points, $m$. To this end, we run sparse GPR as described in \cite{titsias2009variational}, as well as STRIDE using the composition architecture and ACA with rank 50 to approximate the hidden layer. In STRIDE, we now use $S=10$ Markov chains as well as $R=3$ individual updates of the inducing points in each maximisation step \textbf{(M)} only. The size of the candidate set is chosen as $|\mathcal{J}|=1000$. Otherwise, the setup remains the same as in the previous experiment in \Cref{sec:uci}. As the training data we use a random selection of 90\% of the dataset, leaving the remaining 10\% as the test set. We repeat this experiments for $m$ ranging from 25 to 125. The obtained SMSE values are presented as an error bar plot in \Cref{fig:fashion_mnist}, with the exact numbers given in \Cref{tab:fashion}. For each value of $m$, we run the experiment five times, using a different train/test split of the dataset for each individual run. The values shown in \Cref{fig:fashion_mnist} and \Cref{tab:fashion} represent the average over five runs. The error bars in \Cref{fig:fashion_mnist} cover $\pm 2$ standard deviations, while the values given in parentheses in \Cref{tab:fashion} are the obtained standard deviations.

Notably, the reported errors decrease as expected as we increase $m$. Even for small numbers of $m$, we obtain a more accurate regression result when using the deep GP model. However, even for sparse GPR, the errors do not stagnate as we increase $m$, indicating that this dataset requires large numbers of inducing points for an accurate representation.

\begin{figure}
    \begin{floatrow}
        \ffigbox{%
            \includegraphics[height=0.525\linewidth]{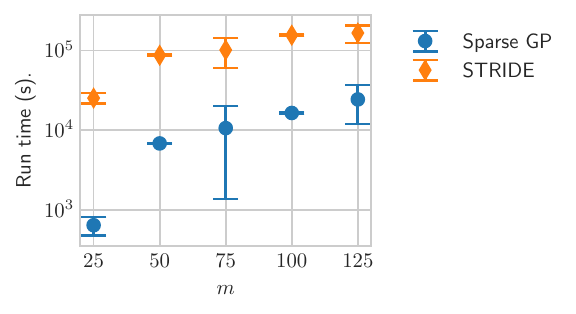}
        }{%
            \caption{Offline timings for varying $m$.}%
            \label{fig:fashion_mnist_t_offline}%
        }
        \quad
        \ffigbox{%
            \includegraphics[height=0.525\linewidth]{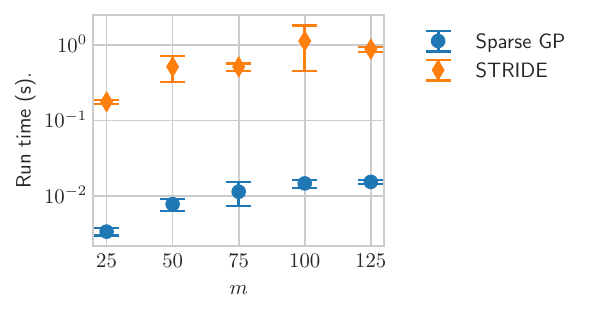}
        }{%
            \caption{Online timings for varying $m$.}%
            \label{fig:fashion_mnist_t_online}%
        }
    \end{floatrow}
\end{figure}

We present the computation timings obtained for this experiment in \Cref{fig:fashion_mnist_t_offline,fig:fashion_mnist_t_online}. They are split up into offline training times in the plot on the left and online inference times in the right plot.
While sparse GP regression is significantly cheaper, computationally speaking, the cost of STRIDE scales similarly to sparse GP regression with increasing $m$. This is as we would expect from our discussion of the computational costs in \Cref{sec:cost}.

\section{Conclusion} \label{Sec_conclusions}
In the above, we have introduced STRIDE, a scalable technique for performing fully Bayesian deep GPR. This method combines the computational savings of sparse GPR with the flexibility of deep GP models and the accurate uncertainty quantification provided by MCMC. We furthermore describe two approaches to approximating the covariance matrices of the hidden layers of the deep GP. In conjunction with STRIDE, this results in a method whose computational cost scales linearly with respect to the number of observations. In experiments on standard datasets, we showed that STRIDE indeed reaches regression accuracies similar to, and sometimes better than, the usually intractable standard GPR, and beats sparse GPR. Thus, STRIDE is a computationally efficient methodology for Bayesian machine learning.

Important future research directions are generalisations of STRIDE to non-Gaussian likelihoods, the SASPA framework might be a useful point to start \cite{Qi}. Moreover, we have not analysed STRIDE in this work, techniques from \cite{finocchio2023posterior,abraham2023deep,Burt,Osborne} might support a convergence analysis.

However, while STRIDE makes deep GPR more scalable, some limitations remain.
Especially the use of STRIDE with high input dimension $d$ without separate feature extraction, for instance, should still be very challenging. STRIDE profits from sparse GPR, but the more flexible deep GP models still require more computational resources.

\begin{ack}
S.U, A.L.T., and J.L.\ acknowledge the use of the HWU high-performance computing facility (DMOG) and associated support services in the completion of this work. A.L.T.\ was partially supported by EPSRC grants EP/X01259X/1 and EP/Y028783/1.
\end{ack}

\bibliographystyle{plainnat}
\bibliography{references}

\end{document}